\documentclass{article}

\PassOptionsToPackage{numbers, compress}{natbib}
\usepackage[preprint]{neurips_2026}

\usepackage[utf8]{inputenc} 
\usepackage[T1]{fontenc}    
\usepackage{hyperref}       
\usepackage{url}            
\usepackage{booktabs}       
\usepackage{amsfonts}       
\usepackage{nicefrac}       
\usepackage{microtype}      
\usepackage{xcolor}         

\usepackage{enumitem}
\usepackage{amsmath}
\usepackage{amssymb}
\usepackage{graphicx} 
\usepackage{multirow}
\usepackage{wrapfig}
\usepackage{threeparttable}
\usepackage[table, dvipsnames]{xcolor}
\newcommand{\secref}[1]{§~\ref{#1}}

\title{Towards Minimal Sufficient Representation for Vision-Language Navigation}

\author{%
\textbf{Yihao Wu}\textsuperscript{1}\thanks{Equal contribution.} \quad
\textbf{Chenyi Xu}\textsuperscript{1}\footnotemark[1] \quad
\textbf{Liqi Yan}\textsuperscript{1}\thanks{Corresponding author.} \quad
\textbf{Chenhuan Cai}\textsuperscript{2} \quad
\textbf{Geyong Min}\textsuperscript{3} \\
\textbf{Bin Lin}\textsuperscript{1} \quad
\textbf{Fangli Guan}\textsuperscript{1} \quad
\textbf{Jianhui Zhang}\textsuperscript{1} \quad
\textbf{Pan Li}\textsuperscript{1} \\
\\[-0.5em]
{\small
\textsuperscript{1}Hangzhou Dianzi University \quad
\textsuperscript{2}University of Zurich \quad
\textsuperscript{3}University of Exeter
} \\
\\[-0.75em]
{\footnotesize
\texttt{wuyihaowyh@gmail.com} \quad
\texttt{chenyixu2004@gmail.com} \quad
\texttt{ylq@hdu.edu.cn}
}
}

\begin{document}
\maketitle

\begin{abstract}
Vision-and-Language Navigation in continuous environments (VLN-CE) requires an agent to ground language in egocentric observations and plan in unseen scenes. Although recent multimodal large models and world-model-based methods have improved Navigation, they often preserve excessive task-irrelevant detail, weakening generalization and increasing computational burden. We propose \textbf{CompactNav}, a navigation framework grounded in the \textit{Principle of Minimal Sufficiency}. CompactNav consists of three components: Logical Anchor Model that implements instruction-aware selective perception to suppress environmental noise, Minimalist Constraint Alignment module that serves as a compact cross-modal bottleneck, efficiently synchronizing discrete linguistic intent with continuous latent dynamics while filtering out redundant information, and Compression World Model that predicts action-conditioned states within a condensed, low-rank latent space. These modules align semantic intent with spatial perception, enhancing the agent's robustness and efficiency in complex tasks. Experiments show that CompactNav improves over prior SOTA by 2.0\%/1.0 in SR/SPL on R2R-CE val-unseen and 0.94\%/0.78 on RxR-CE val-unseen. These results indicate that minimally sufficient world representations provide an effective foundation for robust VLN. \textit{All code will be publicly released.}

\end{abstract}

\section{Introduction}

Embodied intelligence has emerged as a fundamental paradigm in artificial intelligence, aiming to enable agents to perceive, reason, and act autonomously in complex physical environments \cite{liu2025aligning,xu2024survey}. Among embodied benchmarks, Vision-and-Language Navigation in Continuous Environments (VLN-CE) \cite{krantz_vlnce_2020,hong2022bridging} serves as a canonical testbed for evaluating multimodal grounding and long-horizon embodied decision-making. In VLN-CE, agents must follow natural language instructions to navigate toward target locations in previously unseen continuous 3D environments using egocentric RGB-D observations \cite{chang2017matterport3dlearningrgbddata}. Despite recent progress, robust navigation under cluttered scenes and long-horizon trajectories remains challenging \cite{song2025towards}, primarily due to the difficulty of learning task-aligned world representations that bridge symbolic linguistic reasoning and high-dimensional perceptual dynamics.

Recent advances in VLN-CE increasingly emphasize multimodal representation learning and latent world modeling. Conventional World Models (WMs) prioritize environmental fidelity by encoding fine-grained spatial-temporal dynamics into latent states to facilitate prediction and planning. Such representations provide strong scene awareness and support long-range behavioral reasoning in complex environments. Concurrently, recent multimodal architectures employ Q-Former or query-token based paradigms \cite{li2023blip,jaegle2021perceiver,alayrac2022flamingo,zhang2025cross} to compress visual observations and trajectory histories into compact latent tokens, thereby improving computational scalability and cross-modal interaction efficiency. These studies validate the importance of latent abstraction for suppressing perceptual redundancy while strengthening semantic alignment between language instructions and visual observations. Collectively, prior works indicate that effective VLN systems critically depend on jointly optimizing spatial-temporal reasoning, multimodal correspondence, and compact representation learning.

Despite these advances, existing methods still face several fundamental challenges in establishing minimally sufficient world representations for robust navigation. \textbf{\textcircled{\scriptsize{1}} Semantic granularity mismatch causes instruction dilution.} The intrinsic entropy gap between sparse symbolic instructions and dense perceptual observations forces global attention mechanisms to jointly process heterogeneous modalities, causing task-relevant cues to be overwhelmed by visual redundancy \cite{liu2026span}. As a result, logical constraints from instructions are often weakened before entering the world representation. \textbf{\textcircled{\scriptsize{2}} High-entropy representations introduce redundancy-induced overfitting.} Fidelity-oriented world models tend to preserve excessive environmental details, such as textures or illumination variations, that are irrelevant to navigation objectives \cite{peng2026structured}. This redundancy not only increases representational complexity but also reduces generalization ability in visually unfamiliar or monotonous environments. \textbf{\textcircled{\scriptsize{3}} Dynamic coordination between intention and perception remains computationally inefficient.} Existing embodied frameworks often lack an efficient mechanism for synchronizing discrete linguistic intent with continuous perceptual dynamics, leading to substantial computational overhead and unstable cross-modal optimization in high-dimensional latent spaces \cite{liu2025navforesee,lian2026mapdream}. 

To address these issues, we propose \textbf{CompactNav}, a navigation framework grounded in the Principle of Minimal Sufficiency. CompactNav integrates three complementary modules for efficient and robust navigation. \textit{First}, we introduce the Focus-Former, which reformulates multimodal interaction as an instruction-conditioned filtering process. Instead of uniformly attending to all perceptual information, Focus-Former extracts task-essential Logical Anchors under linguistic constraints, allowing visual features to be selectively activated only when they satisfy instruction semantics. \textit{Second}, we design a Minimalist Compression World Model (CWM) that performs spatial-temporal abstraction within a highly condensed latent manifold, enabling the agent to preserve only topology- and geometry-related information necessary for path planning while discarding irrelevant environmental entropy. \textit{Third}, we establish a differentiable low-rank alignment mechanism to realize Minimalist Constraint Alignment (MCA). By factorizing the world tensor into independent low-rank basis vectors, MCA constructs an explicit information bottleneck that efficiently aligns instruction intent with latent world dynamics, while reducing representational complexity from $\mathcal{O}(N\cdot M\cdot D)$ to $\mathcal{O}(R(N+M+D))$.

\begin{figure*}[t!]
\vspace{-8pt}
    \centering
    \includegraphics[width=1\linewidth]
    {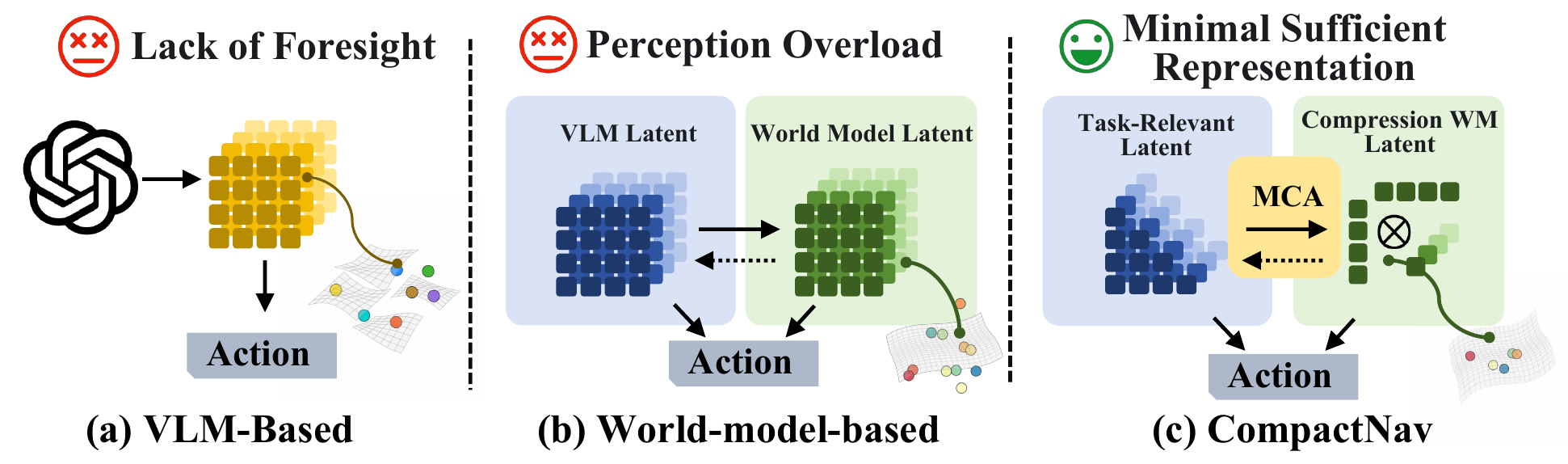}
    \vspace{-10pt}
    \caption{\textbf{Comparison with representative VLN paradigms.} (a) VLM-based policies directly map observations and instructions to actions but lack explicit future modeling. (b) World-model-based frameworks incorporate future-state prediction yet remain limited by perceptual redundancy. (c) CompactNav introduces a low-rank cross-modal alignment mechanism through the proposed MCA module to learn minimally sufficient latent representations for robust embodied navigation.}
    \label{fig:teaser}
\vspace{-15pt}
\end{figure*}

Our contributions are summarized as follows:
\begin{itemize}[leftmargin=*]
\item We introduce CompactNav, a framework that shifts the VLN-CE paradigm from fidelity-maximization to Minimal Sufficiency. Through instruction-conditioned filtering and low-rank state factorization, CompactNav reduces perceptual redundancy by maintaining task-congruent representations while actively discarding irrelevant environmental entropy.
\item We propose a modular architecture with the Focus-Former for logic anchoring and the CWM for compact latent prediction. This design bridges the instruction-observation entropy gap and supports stable planning within a condensed latent manifold.
\item We establish an MCA module based on CP factorization. By factorizing the world state into low-rank basis vectors, we enable the instruction's logical intent to dynamically constrain the manifold evolution, ensuring computational efficiency and robust cross-modal alignment.
\item We develop a training paradigm transitioning from multimodal alignment to end-to-end co-evolution. By distilling the Information Bottleneck (IB) into Latent Manifold Minimization (LMM), we enforce structural sparsity. This two-phase optimization enables the Focus-Former and CWM to jointly evolve for minimal yet sufficient representations in robust navigation.
\end{itemize}

\section{Related Work}
\textbf{Vision-and-Language Navigation in Continuous Environments.}
VLN studies how an embodied agent follows natural language instructions to reach a target location in previously unseen environments. Early benchmarks such as Room-to-Room (R2R) \cite{anderson2018vision} and Room-across-Room (RxR) \cite{ku2020room} were built on discretized Matterport3D navigation graphs \cite{chang2017matterport3dlearningrgbddata}. Recent work has shifted toward continuous environments (VLN-CE) \cite{krantz_vlnce_2020,hong2022bridging}, where partial observability and low-level control make long-horizon grounding and planning substantially more challenging. Existing VLN-CE methods mainly build explicit spatial abstractions, such as waypoint predictors, topological maps, or structured scene representations \cite{hong2022bridging, zhang2025cosmo, gao20253d, an2024etpnav,wang2023dreamwalker,wang2024lookahead}, or leverage large pre-trained vision-language models to infer actions and decompose instructions into executable subgoals \cite{zhou2024navgpt,zhang2024navidvideobasedvlmplans,zhang2025uninavidvideobasedvisionlanguageactionmodel,qi2025vlnr1visionlanguagenavigationreinforcement}. Despite their progress, existing methods still struggle to balance fine-grained instruction grounding, compact environment representation, and efficient long-horizon decision-making in unseen scenes.

\textbf{World Model.}
World models provide embodied agents with a predictive latent space for encoding observations and simulating future state transitions, enabling planning beyond immediate perception \cite{hafner2019dream,hafner2019learning}. In navigation, this idea has evolved from early view-generation methods that synthesize unseen observations for augmentation or short-horizon lookahead \cite{koh2021pathdreamer,li2023panogen, perincherry2025visual} to more structured latent simulators that support predictive rollouts in continuous vision-and-language navigation \cite{wang2023dreamwalker,yao2025navmorph,wang2025dreamnavtrajectorybasedimaginativeframework,bar2025navigation}. These studies demonstrate the value of latent prediction for long-horizon decision making, but existing approaches generally emphasize future-scene generation or fixed latent dynamics, without explicitly imposing an instruction-guided information bottleneck on the world representation. 

\textbf{Low-Rank Modeling.}
Low-rank modeling is a natural strategy for removing redundancy from high-dimensional visual and latent representations, with classical tensor decompositions such as CP \cite{hitchcock1927expression} and Tucker \cite{tucker1966some} providing compact multilinear structure. In modern neural latent modeling, one line learns factorized latent subspaces to separate content \cite{mathieu2016disentanglingfactorsvariationdeep,khan2022hamiltonian}, dynamics \cite{pan2022iso}, or control factors \cite{gao2025adaworld,wang2026factoredlatentactionworld}, improving interpretability, controllability, and generalization. Another line compresses latent features through sparsity, gating, pruning, or information-bottleneck-style regularization to suppress task-irrelevant information and improve efficiency and robustness \cite{dai2018compressing,wieczorek2018learningsparselatentrepresentations,torfi2018attentionbasedguidedstructuredsparsity,sun2024learninglatentdynamicrobust,bai2025rethinkinglatentredundancybehavior,liu2026enhancedstructuredlassopruning}. However, existing methods usually emphasize either structural factorization or task-driven compression in generic latent spaces, rather than combining both in an instruction-conditioned manner.

\section{Methodology}
\begin{figure*}[htb]
\vspace{-12pt}
    \centering
    \includegraphics[width=1\linewidth]{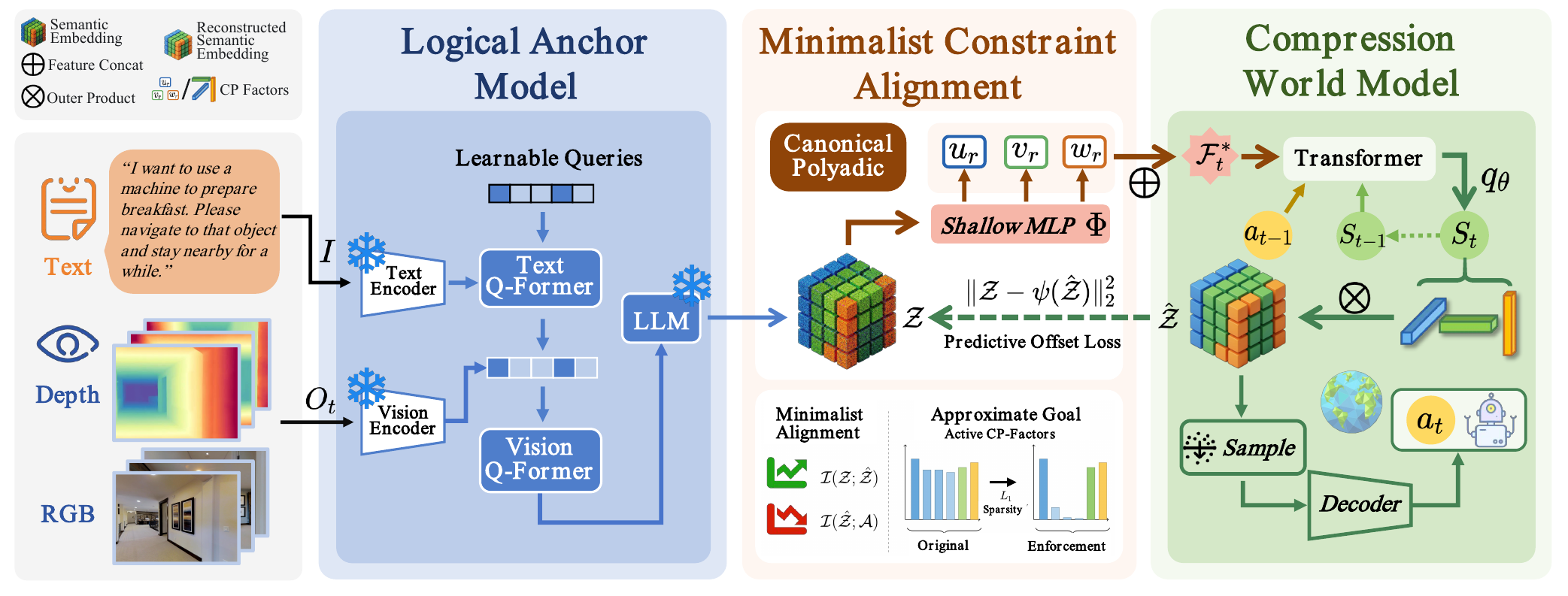}
    \vspace{-17pt}
    \caption{\textbf{Overview of CompactNav.} Given a language instruction together with RGB-D observations, CompactNav first employs the (\secref{sec:LAM}) \textit{\textbf{Logical Anchor Model}} to align instruction semantics with visual perception and produce an instruction-grounded latent embedding $\mathcal{Z}$. This embedding is then compressed through (\secref{sec:MCA}) \textit{\textbf{Minimalist Constraint Alignment}}, where a shallow MLP predicts CP factors $\mathbf{u}$, $\mathbf{v}$, and $\mathbf{w}$ to reconstruct a compact representation $\hat{\mathcal{Z}}$ that preserves task-relevant semantic and structural information while suppressing redundant perceptual content. Based on this compressed representation, the (\secref{sec:CWM}) \textbf{\textit{Compression World Model}} performs action-conditioned latent state transition, maintains a compact predictive belief over future dynamics, and decodes the next navigation action, thereby supporting efficient long-horizon planning.}
\vspace{-8pt}
\end{figure*}


\textbf{Task Definition.} We formulate VLN-CE as a decision-making process within the Habitat simulator. Given a natural language instruction $I$ and an RGB-D observation $O$ from the Matterport3D dataset, the agent aims to reach a target location by predicting a sequence of actions $\mathcal{A}$. At each timestep $t$, the agent perceives an observation $o_t$ and infers an action $a_t$ under a learnable policy $\pi$, where $o_{0:t-1}$ and $a_{0:t-1}$ denote the history of observations and actions, respectively. These actions $a_t \in \mathcal{A}$ are subsequently mapped into low-level control commands for execution. An episode terminates when the agent invokes a stop signal or exceeds the maximum step constraint $T_{max}$.

\subsection{Logical Anchor Model}
\label{sec:LAM}
The core obstacle in VLN lies in the \textit{Semantic Granularity Mismatch}: the instruction $I$, as a structured sequence, exhibits moderate global entropy due to grammatical constraints, while critical tokens (e.g., verbs, objects) within $I$ carry high local information content. Conversely, the observation $O$ consists of pixels with low individual entropy, but its high-dimensional combination results in a high-entropy perceptual space, overwhelming task-relevant signals.

To understand how the Focus-Former addresses this problem, we first revisit the traditional Q-Former. Given an instruction, its feature representation is $F^{(\mathcal{I})} \in \mathbb{R}^{n \times d}$, where $n$ is the number of tokens. Similarly, the observation feature representation is $F^{(\mathcal{O})} \in \mathbb{R}^{m \times d}$ where $m$ is the number of tokens. A conventional approach utilizes learnable queries \cite{li2023blip,jaegle2021perceiver} $Q_A \in \mathbb{R}^{k \times d}$, where $k$ is the number of queries, to perform global attention over the concatenated features $[F^{(\mathcal{I})}, F^{(\mathcal{O})}] \in \mathbb{R}^{(n + m) \times d}$. 

However, this formulation treats tokens from disparate modalities as homogeneous entries in a unified sample space, implicitly performing attention over a joint representation $P(F^{(\mathcal{I})} \cup F^{(\mathcal{O})} \mid Q_A)$. As a result (Figure~\ref{fig:former}), the instruction signals are forced into a competition with the massive observation noise for the limited attention budget:
{
\small
\begin{equation}
\sum \text{Attention}(Q_A, F^{(\mathcal{I})}) + \sum \text{Attention}(Q_A, F^{(\mathcal{O})}) = 1.
\end{equation}
}

In high-noise environments where $|F^{(\mathcal{O})}_{\text{noise}}| \gg |F^{(\mathcal{I})}|$, the instruction signal is weakened, leading to a failure to identify task-critical logical anchors. The output of this conventional approach can be expressed as:
{
\small
\begin{equation}
\label{eq:ha}
\mathbf{H}_{A} = \text{Softmax} \left( \frac{(Q_AW_Q) ([F^{(\mathcal{I})}, F^{(\mathcal{O})}]W_K)^\top}{\sqrt{d}} \right) ([F^{(\mathcal{I})}, F^{(\mathcal{O})}]W_V)
\end{equation}
}

To resolve this, Focus-Former reformulates the interaction as a conditional filtering mechanism \cite{alayrac2022flamingo} $P(F^{(\mathcal{I})} \mid Q_B) \cdot P(F^{(\mathcal{O})} \mid F^{(\mathcal{I})}, Q_B)$, leveraging the low-entropy instruction as a prior to deconstruct the perceptual high-entropy bottleneck. The process unfolds in two topological stages:
\begin{enumerate}[leftmargin=*]
\vspace{-4pt}
    \item \textbf{Logic Anchoring (Instruction Distillation):} The Text Q-former conditions $Q_B$ on $F^{(\mathcal{I})}$. This acts as a logic grounding process in the feature manifold, transforming $Q_B$ into a latent state $Q_B' = \text{Attn}(Q_B, F^{(\mathcal{I})}, F^{(\mathcal{I})})$ that is aligned with the instruction's logic.
    \item \textbf{Perceptual Selection (Task-driven Filtering):} The anchored queries $Q_B'$ then interact with $F^{(\mathcal{O})}$. Crucially, as $Q_B'$ has been pre-conditioned within the instruction's semantic subspace, the subsequent attention over $F^{(\mathcal{O})}$ is no longer a cross-modal competition but a targeted selection \cite{locatello2020object}: $\mathbf{H}_{\text{B}} = \text{Attn}(Q_B', F^{(\mathcal{O})}, F^{(\mathcal{O})})$.
\vspace{-4pt}
\end{enumerate}

Expanding the composite function reveals the structural difference in the attention matrix:
{
\small
\begin{equation}
\mathbf{H}_{B} = \text{Softmax} \left( \frac{ (\text{Attn}(Q_B, F^{(\mathcal{I})}, F^{(\mathcal{I})}) W_Q ) (F^{(\mathcal{O})}W_K)^\top} {\sqrt{d}} \right) (F^{(\mathcal{O})}W_V).
\end{equation}
}

Unlike the global softmax in the joint distribution model (Eq.~\ref{eq:ha}), our architecture enforces a cascaded normalization scheme. The queries $Q_B'$ are pre-conditioned on the linguistic manifold, ensuring they are structurally shielded from irrelevant visual noise before attending to the observation $F^{(\mathcal{O})}$. This formulation ensures that visual features are only activated if they satisfy the logical constraints of the instruction, achieving robust Logic Anchoring even in high-entropy environments.

Following the Focus-Former, the distilled representations $\mathbf{H}_B$ are fed into the MLLM (Qwen2.5-VL-7B~\cite{qwen2025qwen25technicalreport}). The MLLM integrates these features with the instruction features $\mathcal{F}^{(\mathcal{I})}$:
{
\small
\begin{equation}
\mathcal{Z} = \text{MLLM}( [\mathbf{H}_B, \mathcal{F}^{(\mathcal{I})}] ) .
\end{equation}
}

Through its hierarchical Transformer blocks, the MLLM performs high-level reasoning to bridge remaining granularity gaps. By leveraging the MLLM’s extensive prior knowledge, the model ensures that the visual semantics and the linguistic goals contribute to latent world state generation.


\subsection{Minimalist Constraint Alignment}
\label{sec:MCA}

To reduce visual redundancy, MCA establishes a differentiable low-rank alignment channel that enforces a Minimalist Constraint by representing the perceptual stream as a low-rank tensor. This is grounded in the CP Decomposition \cite{kolda2009tensor}, which disentangles multi-modal correlations and resolves the computational bottleneck by approximating the high-order world tensor $\mathcal{Z} \in \mathbb{R}^{N \times M \times D}$ as a sum of rank-one components.

Instead of transmitting the full latent world tensor $\mathcal{Z} \in \mathbb{R}^{N \times M \times D}$, where $N$ represents the number of panoramic segments, $M$ represents the number of tokens, and $D$ represents the feature dimension, we decompose it into a latent space governed by the CP Decomposition. We approximate the world state as the sum of $\mathbf{R}$ rank-one tensors. Formally, the latent representation is reconstructed from three basis vectors: $\mathbf{u}_r$, $\mathbf{v}_r$, $\mathbf{w}_r$. 

Rather than optimizing these vectors as static variables, we utilize a shallow MLP $\Psi$ to generate these factors. The MLP processes the raw latent feature $\mathcal{Z}$ to output the factorized basis vectors:
\small
\begin{equation}
\{\mathbf{u}_r, \mathbf{v}_r, \mathbf{w}_r \}_{r=1}^R = \Psi(\mathcal{Z}),
\end{equation}
Specifically, $\Psi$ employs a shared backbone followed by three independent heads to project the joint representation into disentangled factor groups $\mathbf{u} \in \mathbb{R}^{R \times N}$, $\mathbf{v} \in \mathbb{R}^{R \times M}$, and $\mathbf{w} \in \mathbb{R}^{R \times D}$. The reconstructed semantic embedding $\hat{\mathcal{Z}}$ is then computed via the outer product:
\small
\begin{equation}
\hat{\mathcal{Z}} = \sum_{r=1}^{R} \mathbf{u}_r \otimes \mathbf{v}_r \otimes \mathbf{w}_r
\end{equation}
This formulation serves as a "soft" pruning mechanism \cite{kolda2009tensor,novikov2015tensorizing}, where the rank $R$ acts as a hyperparameter controlling the bottleneck capacity. By limiting $R$, we explicitly enforce a low-rank structural constraint on the latent space. Crucially, as the reconstruction $\hat{\mathcal{Z}}$ is fully differentiable, the Navigation Agent can steer the CWM's perception through gradient flow: the MLP $\Psi$ learns to suppress irrelevant features by assigning near-zero magnitudes to the corresponding factor components $\{\mathbf{u}, \mathbf{v}, \mathbf{w}\}$. In this sense, $\Psi$ functions as a task-driven filter, leveraging logical priors from the LAM to distill the high-entropy perceptual stream into a sequence of task-essential anchors.

\textbf{Minimalist Alignment.} This formulation follows the Information Bottleneck (IB) principle \cite{alemi2016deep,tishby2000information}, by enforcing the low-rank constraint $R$, we effectively bound the mutual information $\mathcal{I}(\mathcal{Z}; \hat{\mathcal{Z}})$, acting as a compressive bottleneck that filters out redundant perceptual noise. Simultaneously, we optimize $\Psi$ to maximize the task-relevant information $\mathcal{I}(\hat{\mathcal{Z}}; \mathcal{A})$, where $\mathcal{A}$ denotes the target action sequence. This ensures that the compressed representation $\hat{\mathcal{Z}}$ preserves only the essential sparse cues necessary for waypoint prediction, reducing perceptual redundancy while maintaining high predictive sufficiency. In implementation, the CP reconstruction is deterministic; the IB notation is used to describe the intended capacity-controlled bottleneck, and the sparsity regularizer below provides its practical surrogate.

\subsection{Compression World Model}
\label{sec:CWM}
To adhere to the Principle of Minimal Sufficiency, we introduce the CWM for continuous spatial-temporal abstraction over the compact representation produced by LAM and MCA. Unlike traditional world models that struggle with high-dimensional pixel reconstruction, the CWM maintains environment dynamics within the condensed latent manifold defined by CP factors. This formulation explicitly filters out transient visual noise, as stochastic perturbations typically manifest as high-rank residuals that fall outside the capacity of our low-rank manifold.

We define the agent’s internal belief state as a stochastic latent variable $\mathbf{s}_t$. Crucially, our architecture achieves low-bandwidth state prediction by operating exclusively within the factorized latent space. Given the partial observability of the environment, we denote the set of rank-one basis vectors as $\mathcal{F} = \{ \mathbf{u}_r, \mathbf{v}_r, \mathbf{w}_r \}$. The CWM performs action-conditioned state transitions to simulate the environmental response to the agent's decisions. Beyond merely encoding observations, the CWM serves as a \textit{Latent Simulator} that anticipates future states without immediate sensory input. This forward prediction is governed by a Prior Transition Network, which learns the temporal evolution of the environment within the factorized vectors. Formally, for each timestep $t$, the CWM maintains both a posterior distribution $q_\phi$ (incorporating the current latent world $\mathcal{F}_t$) and a prior distribution $p_\theta$ (predicting from history). The posterior acts as an observation-conditioned belief update during training, while the prior learns to reproduce this belief using only previous latent states and actions, making it suitable for future rollout at test time:
\begin{align}
\text{Posterior:} \quad & q_\phi(\mathbf{s}_t \mid \mathcal{F}_{\le t}, \mathbf{a}_{<t}) \
\sim \mathcal{N}(\mu_\phi(\mathcal{F}_t, \mathbf{a}_{t-1}, \mathbf{s}_{t-1}), \sigma_\phi(\mathcal{F}_t, \mathbf{a}_{t-1}, \mathbf{s}_{t-1}){E}), \\
\text{Prior:} \quad & p_\theta(\mathbf{s}_t \mid \hat{\mathbf{s}}_{t-1}) \sim \mathcal{N}(\hat{\mu}_\theta(\hat{\mathbf{s}}_{t-1}, \hat{\mathbf{a}}_{t-1}), \sigma_\theta (\hat{\mathbf{s}}_{t-1}, \hat{\mathbf{a}}_{t-1})E),
\end{align}
where $\mathbf{s}_t$ is the latent state, $\mathbf{a}_{t-1}$ is the previous action, and $\mathcal{F}_t$ is the task-relevant factors refined via minimalist alignment, where $\hat{\mathbf{a}}_{t-1}$ denotes actions sampled during prediction. By conditioning the dynamics on the joint representation of previous states and CP-distilled factors $\mathcal{F}_t$, the CWM captures the causal interplay between control commands and spatial transformations while remaining computationally lean. Rather than evolving a dense volumetric state $\mathcal{Z} \in \mathbb{R}^{N \times M \times D}$, our model operates exclusively within the factorized manifold, reducing the representational complexity from $\mathcal{O}(N\cdot M\cdot D)$ to $\mathcal{O}(R(N+M+D))$. This factorized latent evolution functions as a task-driven filter, effectively bounding accumulated errors through LAM-provided logical anchors. Such a reduction in information throughput enables the agent to perform topologically consistent prediction of future spatial layouts, ensuring long-horizon predictive sufficiency with minimal computational overhead.

\subsection{Synergistic Training: From Alignment to Co-evolution}
\label{sec:TO}
The optimization of CompactNav is formulated as a multi-objective task, aiming to maximize task-relevant information within a compressed latent manifold while ensuring the consistency of environmental dynamics. The total loss is: $\mathcal{L} =  \mathcal{L}_{WM} + \lambda_1 \mathcal{L}_{LMM} + \lambda_2 \mathcal{L}_{PO} + \mathcal{L}_{IL},$ where $\lambda_1$ and $\lambda_2$ denote the weighting coefficients.

\subsubsection{Phase I: Multimodal Alignment Initialization} In the initial phase, we establish a communication channel between the LAM and CWM to synchronize feature extraction with latent dynamics. The agent first observes inputs over $T$ timesteps and subsequently predicts latent trajectories for an additional $T_{ext}$ steps.

\textbf{Compression World Model Initialization.} Simultaneously, the CWM is trained to model environmental transitions via a self-supervised variational objective. We minimize the KL divergence between the posterior belief $q$ and the prior prediction $p$:
\small
\begin{equation}
\begin{aligned}
\mathcal{L}_{WM} = \sum_{t=1}^{T+T_{ext}}{\mathbb{E}_q\left[ \underbrace{\log p(\mathcal{F}_t \mid s_t)}_{\text{Reconstruction}} + \underbrace{\log p(a_t \mid s_t)}_{\text{Action Prediction}} \right]} 
- \sum_{t=1}^{T}\underbrace{D_{KL}\bigl(q(s_t \mid \mathcal{F}_{\le t}, a_{< t}) \,\|\, p(s_t \mid s_{t-1})\bigr)}_{\text{Posterior and Prior Match}}.
\end{aligned}
\end{equation}

\textbf{Multimodal Bottleneck Alignment.} To enforce minimality while preserving task-relevant information, we optimize the Focus-Former using an Information Bottleneck (IB) objective. Since $\hat{\mathcal{Z}}$ is generated through CP factors, the variational form below is instantiated as a sparsity-inducing penalty on those factors rather than as an additional sampling module:
{
\small
\begin{equation}
\label{eq:ib_laplace}
\mathcal{L}_{\text{IB}} = \underbrace{I(\mathcal{Z}; \hat{\mathcal{Z}})}_{\text{Minimality}} - \beta \underbrace{I(\hat{\mathcal{Z}}; \mathcal{A})}_{\text{Sufficiency}}
\approx \mathbb{E}_{p(\mathcal{Z})} \left[ D_{KL}\bigl(q(\hat{\mathcal{Z}} \mid \mathcal{Z}) \,\|\, p(\hat{\mathcal{Z}})\bigr) \right] - \beta \mathbb{E}_{q(\hat{\mathcal{Z}} \mid \mathcal{Z})} \left[ \log p(\mathcal{A} \mid \hat{\mathcal{Z}}) \right],
\end{equation}
}

where $p(\hat{\mathcal{Z}})$ is a factorized Laplace prior. The $\beta$ balances minimality and sufficiency. Assuming a fixed-variance variational posterior, the KL term simplifies to an $L_1$-regularized penalty on the CP factors:
\small
\begin{equation}
\label{eq:kl_laplace}
\mathcal{L}_{\text{LMM}} = \mathbb{E}_{p(\mathcal{Z})} \left[ D_{KL}\bigl(q(\hat{\mathcal{Z}} \mid \mathcal{Z}) \,\|\, p(\hat{\mathcal{Z}})\bigr) \right] \propto \sum_{r=1}^R \left( \|\mathbf{u}_r\|_1 + \|\mathbf{v}_r\|_1 + \|\mathbf{w}_r\|_1 \right) + \text{const.},
\end{equation}
This Latent Manifold Minimization ($\mathcal{L}_{\text{LMM}}$) loss enforces structural sparsity, effectively truncating redundant latent ranks and distilling the perceptual stream into a minimal set of task-essential basis vectors.

\begin{table*}[t!] 
\vspace{-8pt}
\centering
	\caption{\textbf{Results on R2R-CE dataset.} $^\dagger$ indicates that this model is based on another model in the same group as its backbone. Best results for the panoramic settings are each highlighted in \textbf{bold}.
    }
    \vspace{1pt}
	\label{tab:r2r-ce}
	\resizebox{0.99\textwidth}{!}{
		\begin{threeparttable}
		\begin{tabular}{@{}l|ccccc|ccccc|ccccc@{}}
			\toprule
			  \multicolumn{1}{c|}{} & \multicolumn{5}{c|}{\textbf{Val Seen}} & \multicolumn{5}{c|}{\textbf{Val Unseen}} & \multicolumn{5}{c}{\textbf{Test Unseen}} \\
			  \multicolumn{1}{c|}{\multirow{-2}{*}{\textbf{Methods}}} & TL & \cellcolor{red!25}NE $\downarrow$ & \cellcolor{gray!25}OSR $\uparrow$ & \cellcolor{gray!25}SR $\uparrow$ & \cellcolor{gray!25}SPL $\uparrow$ & TL & \cellcolor{red!25}NE $\downarrow$ & \cellcolor{gray!25}OSR $\uparrow$ & \cellcolor{gray!25}SR $\uparrow$ & \cellcolor{gray!25}SPL $\uparrow$ & TL & \cellcolor{red!25}NE $\downarrow$ & \cellcolor{gray!25}OSR $\uparrow$ & \cellcolor{gray!25}SR $\uparrow$ & \cellcolor{gray!25}SPL $\uparrow$ \\ \midrule
			Seq2Seq~\cite{anderson2018vision} {\scriptsize {\color{Gray} [CVPR18]} } & 9.26 & 7.12 & 46 & 37 & \multicolumn{1}{c|}{35} & 8.64 & 7.37 & 40 & 32 & \multicolumn{1}{c|}{30} & 8.85 & 7.91 & 36 & 28 & 25 \\
			Sim2Sim~\cite{Krantz2022Sim2SimTF} {\scriptsize {\color{Gray} [ECCV22]} }& 11.18 & 4.67 & 61 & 52 & \multicolumn{1}{c|}{44} & 10.69 & 6.07 & 52 & 43 & \multicolumn{1}{c|}{36} & 11.43 & 6.17 & 52 & 44 & 37 \\
			CWP-CMA~\cite{Hong2022BridgingTG} {\scriptsize {\color{Gray} [CVPR22]}}& 11.47 & 5.20 & 61 & 51 & \multicolumn{1}{c|}{45} & 10.90 & 6.20 & 52 & 41 & \multicolumn{1}{c|}{36} & 11.85 & 6.30 & 49 & 38 & 33 \\
			CWP-BERT~\cite{Hong2022BridgingTG}{\scriptsize {\color{Gray} [CVPR22]}}& 12.50 & 5.02 & 59 & 50 & \multicolumn{1}{c|}{44} & 12.23 & 5.74 & 53 & 44 & \multicolumn{1}{c|}{39} & 13.51 & 5.89 & 51 & 42 & 36 \\
			DREAMW~\cite{Wang2023DREAMWALKERMP} {\scriptsize {\color{Gray} [ICCV23]}}& 11.60 & 4.09 & 59 & 66 & \multicolumn{1}{c|}{48} & 11.30 & 5.53 & 49 & {59} & \multicolumn{1}{c|}{44} & 11.80 & 5.48 & 49 & {57} & 44 \\
			GridMM~\cite{wang2023gridmm} {\scriptsize {\color{Gray} [ICCV23]}}& 12.69 & 4.21 & 69 & 59 & \multicolumn{1}{c|}{51} & 13.36 & 5.11 & 61 & 49 & \multicolumn{1}{c|}{41} & 13.31 & 5.64 & 56 & 46 & 39 \\
			BEVBert~\cite{an2023bevbert} {\scriptsize {\color{Gray} [ICCV23]}}& 13.98 & 3.77 & 73 & 68 & \multicolumn{1}{c|}{60} & 13.27 & 4.57 & 67 & 59 & \multicolumn{1}{c|}{50} & 15.31 & 4.70 & 67 & 59 & 50 \\ 
			FSTTA~\cite{gao2024fast} {\scriptsize {\color{Gray} [ICML24]}}& 12.39 & 4.25 & 69 & 58 & \multicolumn{1}{c|}{50} & 11.58 & 5.27 & 58 & 48 & \multicolumn{1}{c|}{42} & 13.17 & 5.84 & 55 & 46 & 38 \\ \cmidrule(l){1-16} 
			ETPNav~\cite{an2024etpnav} {\scriptsize {\color{Gray} [TPAMI24]}}& 11.78 & 3.95 & 72 & 66 & \multicolumn{1}{c|}{59} & 11.99 & 4.71 & 65 & 57 & \multicolumn{1}{c|}{49} & 12.87 & 5.12 & 63 & 55 & 48 \\ 
			NavMorph$^\dagger$~\cite{yao2025navmorph}{\scriptsize {\color{Gray} [ICCV25]}} & \multicolumn{1}{c}{11.43} &  \multicolumn{1}{c}{{{{3.86}}}} &  \multicolumn{1}{c}{{{73}}} &  \multicolumn{1}{c}{{{67}}} &  \multicolumn{1}{c|}{{{60}}} & \multicolumn{1}{c}{11.55} & \multicolumn{1}{c}{{{4.62}}} & \multicolumn{1}{c}{{{66}}} & \multicolumn{1}{c}{{{59}}} & \multicolumn{1}{c|}{{{50}}} & 12.88 & {{4.91}} & {{64}} & {{57}} & {{49}}\\ 
            \cmidrule(l){1-16}
			HNR~\cite{wang2024lookahead} {\scriptsize {\color{Gray} [CVPR24]}} & 11.79 & 3.67 & 76 & 69 & \multicolumn{1}{c|}{61} & 12.64 & 4.42 & 67 & 61 & \multicolumn{1}{c|}{51} & 13.03 & 4.81 & 67 & 58 & 50 \\
              \cellcolor{gray!10}NavMorph$^\dagger$~\cite{yao2025navmorph}{\scriptsize {\color{Gray} [ICCV25]}} & \multicolumn{1}{c}{\cellcolor{gray!10}{11.76}} &  \multicolumn{1}{c}{\cellcolor{gray!10}{3.66}} &  \multicolumn{1}{c}{\cellcolor{gray!10}{78}} &  \multicolumn{1}{c}{\cellcolor{gray!10}{70}} &  \multicolumn{1}{c|}{\cellcolor{gray!10}{62}} & \multicolumn{1}{c}{\cellcolor{gray!10}{12.68}} & \multicolumn{1}{c}{\cellcolor{gray!10}{4.37}} & \multicolumn{1}{c}{\cellcolor{gray!10}{68}} & \multicolumn{1}{c}{\cellcolor{gray!10}{64}} & \multicolumn{1}{c|}{\cellcolor{gray!10}{53}} & \cellcolor{gray!10}{12.69} & \cellcolor{gray!10}{4.69} & \cellcolor{gray!10}{68} & \cellcolor{gray!10}{60} & \cellcolor{gray!10}{52}\\
              \cmidrule(l){1-16}
              \cellcolor{gray!25}\textbf{CompactNav (Ours)} & \multicolumn{1}{c}{\cellcolor{gray!25}{11.58}} &  \multicolumn{1}{c}{\cellcolor{gray!25}{{\textbf{3.57}}}} &  \multicolumn{1}{c}{\cellcolor{gray!25}{\textbf{80}}} &  \multicolumn{1}{c}{\cellcolor{gray!25}{\textbf{71}}} &  \multicolumn{1}{c|}{\cellcolor{gray!25}{\textbf{63}}} & \multicolumn{1}{c}{\cellcolor{gray!25}{12.42}} & \multicolumn{1}{c}{\cellcolor{gray!25}{\textbf{4.36}}} & \multicolumn{1}{c}{\cellcolor{gray!25}{\textbf{70}}} & \multicolumn{1}{c}{\cellcolor{gray!25}{\textbf{66}}} & \multicolumn{1}{c|}{\cellcolor{gray!25}{\textbf{54}}} & \cellcolor{gray!25}{12.56} & \cellcolor{gray!25}{\textbf{4.58}} & \cellcolor{gray!25}{\textbf{70}} & \cellcolor{gray!25}{\textbf{63}} & \cellcolor{gray!25}{\textbf{55}}\\
			\bottomrule
		\end{tabular}
		\end{threeparttable} 	
	}\vspace{-8pt}
\end{table*}


\subsubsection{Phase II: End-to-End Policy and Architecture Co-evolution} Upon establishing a task-consistent latent manifold, we introduce the Navigation Agent for joint optimization. In this phase, task-specific gradients propagate from the navigation objective back through the policy to the Focus-Former and CWM, enabling the entire framework to self-evolve.

Since the factorized basis vectors $\{\mathbf{u, v, w}\}$ generated by the MLP $\Psi$ may not perfectly approximate the world tensor, we introduce an auxiliary predictive offset loss to calibrate the low-rank reconstruction $\mathcal{L}_{PO} = \| \mathcal{Z} - \hat{\mathcal{Z}} \|_2^2$. This loss acts as a regularizer, steering the Focus-Former to refine its perceptual filtering by penalizing the divergence between the MLLM's latent $\mathcal{Z}$ and its low-rank counterpart $\hat{\mathcal{Z}}$. The agent utilizes the latent state $s_t$ to predict action $a_t$. To achieve robust decision-making, we append an imitation learning objective: $\mathcal{L}_{IL} = - \sum_{t=1}^{T} \log p(a^*_t \mid I, o_t)$, where $a^*_t$ represents the teacher action at each step.


\begin{wraptable}{r}{0.45\textwidth}
\vspace{-20pt}
    \caption{\textbf{Results on RxR-CE dataset.} This table reports VLN-CE metrics under Val Unseen split of RxR-CE dataset.}
    \vspace{2pt}
    \label{tab:rxr-ce}
    \centering
    \resizebox{\linewidth}{!}{
        \begin{tabular}{@{}l|ccccc@{}}
            \toprule
            \multicolumn{1}{c|}{} & \multicolumn{5}{c}{\textbf{Val Unseen}} \\ 
              \multicolumn{1}{c|}{\multirow{-2}{*}{\textbf{Methods}}} & \cellcolor{red!25}NE $\downarrow$ & \cellcolor{gray!25}SR $\uparrow$ & \cellcolor{gray!25}SPL $\uparrow$ & \cellcolor{gray!25}NDTW $\uparrow$ & \cellcolor{gray!25}SDTW $\uparrow$ \\ 
            \midrule
            LAW-Pano~\cite{Raychaudhuri2021LanguageAlignedW} & 11.04 & 10.0 & 9.0 & - & - \\
            Seq2Seq~\cite{anderson2018vision} & 12.1 & 13.93 & 11.96 & 30.86 & 11.01 \\
            CWP-CMA~\cite{Hong2022BridgingTG} & 8.76 & 26.59 & 22.16 & 47.05 & - \\
            CWP-BERT~\cite{Hong2022BridgingTG} & 8.98 & 27.08 & 22.65 & 46.71 & - \\
            AO-Planner~\cite{chen2024affordances} & 7.06 & 43.3 & 30.5 & 50.1 & - \\
            Reborn~\cite{an20221st} & 5.98 & 48.60 & 42.05 & 63.35 & 41.82 \\
            \cmidrule(l){1-6} 
            ETPNav~\cite{an2024etpnav} & {5.64} & 54.79 & 44.89 & 61.90 & 45.33 \\
            \cellcolor{gray!10}NavMorph~\cite{yao2025navmorph} & \cellcolor{gray!10}{5.80}  & \cellcolor{gray!10}{56.23}  & \cellcolor{gray!10}{46.39}  & \cellcolor{gray!10}{63.23}  & \cellcolor{gray!10}{46.98}  \\
            \cmidrule(l){1-6} 
            HNR~\cite{wang2024lookahead} & \textbf{5.51} & 56.39 & 46.73 & 63.56 & 47.24 \\ 
            NavMorph~\cite{yao2025navmorph} & {5.70}  & 58.02  & 48.98  & 64.77  & 48.85  \\
            \cmidrule(l){1-6} 
            \cellcolor{gray!25}CompactNav & \cellcolor{gray!25}{5.74}  & \cellcolor{gray!25}\textbf{58.96}  & \cellcolor{gray!25}\textbf{49.76} & \cellcolor{gray!25}\textbf{65.67} & \cellcolor{gray!25}\textbf{49.94}  \\
            \bottomrule
        \end{tabular}}
\vspace{-8pt}
\end{wraptable}

\section{Experiments}

\subsection{Experimental Setup}
\textbf{Datasets \& Evaluation Metrics.} We evaluate CompactNav on the two standard VLN-CE benchmarks, \textbf{R2R-CE} and \textbf{RxR-CE}, both instantiated in Habitat on Matterport3D scenes \cite{anderson2018vision,ku2020room,krantz_vlnce_2020}. Following standard practice, we report results on \emph{Val-Seen} and \emph{Val-Unseen}. Compared to R2R-CE, RxR-CE is larger and more challenging, with longer trajectories, longer instructions, and multilingual supervision, making it a stronger test of long-horizon grounding and generalization \cite{ku2020room}. We report the standard VLN-CE metrics: Trajectory Length (TL), Navigation Error (NE), Oracle Success Rate (OSR), Success Rate (SR), Success weighted by Path Length (SPL), Normalized Dynamic Time Warping (NDTW), and Success weighted by Dynamic Time Warping (SDTW) \cite{ilharco2019general}. Together, these metrics reflect navigation accuracy, path efficiency, and trajectory fidelity.

\textbf{Baselines.} To evaluate the effectiveness of the proposed method, we conduct extensive comparative experiments across several state-of-the-art baselines, including ETPNav \cite{an2024etpnav}, HNR \cite{wang2024lookahead}, BEVBert \cite{an2023bevbert}, and NavMorph \cite{yao2025navmorph}. We used two different versions of NavMorph, ETPNav and HNR, as the backbone models respectively. Specifically, ETPNav represents a cornerstone approach for long-range planning in continuous environments, while HNR incorporates a hierarchical reasoning framework to optimize path planning. BEVBert focuses on enhancing spatial awareness through BEV representations. Furthermore, we evaluate our method on NavMorph, a framework that employs a world model as its core navigation component. All experiments are performed on the R2R-CE and RxR-CE benchmarks to ensure a comprehensive and fair evaluation.

\subsection{Comparison with State-of-the-Art (SOTA) Methods}

\textbf{Performance on R2R-CE.} Table~\ref{tab:r2r-ce} demonstrates that CompactNav outperforms existing methods across all splits, achieving 63\% SR and 55 SPL on Test Unseen. Even when compared against the recent baseline, NavMorph, CompactNav exhibits a superior target-localization capability, reducing the NE to 4.36m on the Val Unseen. These gains are attributed to our Focus-Former, which employs a low-rank CP bottleneck to filter out environment-specific entropy. This mechanism limits visual-noise interference, allowing the agent to concentrate on task-critical landmarks. Furthermore, unlike HNR, CompactNav maintains a lower TL (12.56m) while achieving higher success rates. This underscores that our model avoids the perceptual redundancy typical of reconstruction-based methods, yielding more efficient navigation paths in novel environments.

\textbf{Performance on RxR-CE.} Table~\ref{tab:rxr-ce} evaluates CompactNav on the more complex RxR-CE dataset, which demands high fidelity to long instructions. CompactNav achieves the SOTA SR (58.96\%) and SPL (49.76), while significantly enhancing path-following accuracy (65.67 NDTW). This superior alignment between execution and instruction is facilitated by our CWM. Unlike traditional models that struggle with the high-dimensional visual streams of long-horizon tasks, our recursive memory maintains a compact history in a low-dimensional manifold. This allows the agent to internally simulate and verify the path ahead without being confounded by redundant perceptual details. These results confirm that CompactNav balances long-range reasoning with efficient state representation, demonstrating high robustness under the strict constraints of RxR-CE.

\textbf{Inference Efficiency.} Although integrating both the MLLM and the world model typically introduces non-negligible computational overhead, CompactNav maintains highly competitive inference efficiency by optimizing the architectural synergy between perception and reasoning. As shown in Table~\ref{tab:online}, CompactNav achieves SOTA navigation performance while maintaining a highly competitive execution time (236s), which is comparable to the baseline ETPNav-NavMorph (227s), despite the latter using a much smaller model backbone. This efficiency stems from our CWM, which mitigates the latency bottleneck of MLLM inference by streamlining the state-update process. Unlike traditional world models that operate on high-dimensional pixel-level reconstructions, our CWM functions within a factorized latent space optimized via CP decomposition. This design ensures that the agent consumes minimal resources for environmental state transitions and memory maintenance, effectively reserving the bulk of the computational budget for the MLLM’s high-level semantic reasoning.

\begin{table}[tb]
\centering
\begin{minipage}{0.48\textwidth}
\centering
\caption{\textbf{Comparison with SOTA methods under online VLN settings.} $^\dagger$ denotes average execution time per instruction.}
\label{tab:online}
\resizebox{0.96\linewidth}{!}{
    \begin{threeparttable} 
        \begin{tabular}{@{}c|ccccc|c@{}}
            \toprule
            \multirow{2}{*}{\textbf{Methods}} & \multicolumn{5}{c|}{\textbf{R2R-CE Val Unseen}} & \multirow{2}{*}{\begin{tabular}[c]{@{}c@{}}Time$^\dagger$\\ (s)\end{tabular}} \\
            & TL & \cellcolor{red!25}NE $\downarrow$ & \cellcolor{gray!25}OSR $\uparrow$ & \cellcolor{gray!25}SR $\uparrow$ & \cellcolor{gray!25}SPL $\uparrow$ &  \\ 
            \midrule
            ETPNav-FSTTA~\cite{gao2024fast} & 11.58 & 5.27 & 58 & 48 & 42 & 291 \\
            ETPNav-NavMorph~\cite{yao2025navmorph} & 11.55 & 4.62 & 66 & 59 & 50 & \textbf{227} \\ 
            \midrule
            \rowcolor{gray!25}\textbf{CompactNav} & 11.48 & \textbf{4.55} & \textbf{68} & \textbf{60} & \textbf{51} & 236 \\
            \bottomrule
        \end{tabular}         
    \end{threeparttable} 
}
\end{minipage}
\hfill
\begin{minipage}{0.49\textwidth}
\centering
\caption{\textbf{Ablation of core components.} This table reports results on the R2R-CE dataset.}
\label{tab:part}
\resizebox{0.99\linewidth}{!}{
    \begin{threeparttable} 
        \begin{tabular}{ccc|ccccc|c@{}}
            \toprule
             \multicolumn{3}{c|}{{\textbf{CompactNav}}} & \multicolumn{5}{c|}{\textbf{R2R-CE Val Unseen}} & \multirow{2}{*}{\begin{tabular}[c]{@{}c@{}}Time$^\dagger$\\ (s)\end{tabular}} \\ 
            \cmidrule(lr){1-3}
            \textbf{LAM} & \textbf{MCA} & \textbf{CWM} & TL & \cellcolor{red!25}NE $\downarrow$ & \cellcolor{gray!25}OSR $\uparrow$ & \cellcolor{gray!25}SR $\uparrow$ & \cellcolor{gray!25}SPL $\uparrow$ \\
            \midrule
            \cellcolor{gray!25}\checkmark & \cellcolor{gray!25}\checkmark & \cellcolor{gray!25}\checkmark & \cellcolor{gray!25} 12.42 & \cellcolor{gray!25} \textbf{4.36} & \cellcolor{gray!25} \textbf{70} & \cellcolor{gray!25} \textbf{66} & \cellcolor{gray!25} \textbf{54} & \cellcolor{gray!25} \textbf{236} \\
            \midrule
              $\times$ & \checkmark & \checkmark & 13.62 & 4.85 & 68 & 65 & 52 & 252 \\
            \checkmark & $\times$ & $\times$ & 12.48 & 4.48 & 69 & 65 & 53 & 516 \\ 
            \checkmark & \checkmark & $\times$ & 12.80 & 4.66 & 68 & 64 & 52 & 368 \\
            \bottomrule
        \end{tabular}
    \end{threeparttable}
}
\end{minipage}
\vspace{-15pt}
\end{table}


\subsection{Ablation Study}
\subsubsection{Impact of Core Components}

We analyze the contribution of each module in CompactNav on the R2R-CE Val Unseen split (Table~\ref{tab:part}). The full model achieves the best accuracy-efficiency trade-off, reaching 66\% SR and 54 SPL with 236s inference time.

\textbf{Logical Anchor Model (LAM).}
Replacing the Focus-Former with a standard Q-Former degrades performance (SR: $66\%\rightarrow65\%$, NE: $4.36\rightarrow4.85$). This indicates that joint vision-language attention introduces interference under cluttered observations, whereas instruction-conditioned filtering improves landmark grounding.

\textbf{Minimalist Constraint Alignment (MCA).}
Removing MCA significantly increases runtime (236s $\rightarrow$ 516s) and reduces navigation performance. This indicates that CP-based factorization is essential not only for compression, but also for enabling low-bandwidth interaction between semantic reasoning and latent dynamics.

\textbf{Compression World Model (CWM).}
Replacing CWM with a standard world model further increases latency (236s $\rightarrow$ 368s) and reduces SR (66\% $\rightarrow$ 64\%). In contrast to reconstruction-heavy models, the proposed CWM operates in a low-rank latent manifold, supporting efficient and topologically consistent prediction for long-horizon navigation.

\subsubsection{Impact of Synergistic Training}
\begin{wraptable}{r}{0.43\textwidth}
\vspace{-20pt}
\centering
\caption{\textbf{Ablation study on loss components.} This table analyzes the contribution of each loss under synergistic training.}
\label{tab:ablation}
\vspace{2pt}
\resizebox{0.99\linewidth}{!}{
\begin{threeparttable} 
\begin{tabular}{@{}l|ccccc@{}}
    \toprule
    \multicolumn{1}{c|}{\multirow{2}{*}{\textbf{Methods}}} &  \multicolumn{5}{c}{\textbf{R2R-CE Val Unseen}} \\ 
    \multicolumn{1}{c|}{} & TL & \cellcolor{red!25}NE $\downarrow$ & \cellcolor{gray!25}OSR $\uparrow$ & \cellcolor{gray!25}SR $\uparrow$ & \cellcolor{gray!25}SPL $\uparrow$  \\ \midrule
    \multicolumn{1}{l|}{\cellcolor{gray!25}{\textbf{CompactNav}}} &  \cellcolor{gray!25}{12.42} & \cellcolor{gray!25}{\textbf{4.36}} & \cellcolor{gray!25}{\textbf{70}} & \cellcolor{gray!25}{\textbf{66}} & \cellcolor{gray!25}{\textbf{54}} \\
    \midrule
    \multicolumn{1}{l|}{\phantom{$\mathcal{L}_{WM}$} $|$ \textit{w/o} $\mathcal{L}_{Re}$} & 11.87 & 4.54 & 68 & 64 & 53 \\
    \multicolumn{1}{l|}{$\mathcal{L}_{WM}$ $|$ \textit{w/o} $\mathcal{L}_{AP}$} & 16.24 & 4.88 & 69 & 63 & 52 \\
    \multicolumn{1}{l|}{\phantom{$\mathcal{L}_{WM}$} $|$ \textit{w/o} $\mathcal{L}_{PPM}$} & 15.92 & 5.12 & 69 & 61 & 50 \\
    \midrule
    \multicolumn{1}{l|}{$\quad \quad$ \textit{w/o} $\mathcal{L}_{LMM}$} & 13.01 & 4.49 & 68 & 64 & 53 \\
    \multicolumn{1}{l|}{$\quad \quad$ \textit{w/o} $\mathcal{L}_{PO}$} & 14.74 & 4.77 & 67 & 63 & 51 \\ 
    \bottomrule
\end{tabular}          
\end{threeparttable}
}\vspace{-8pt}
\end{wraptable}
To validate the design of the CWM, we ablate its core loss components in Table~\ref{tab:ablation}. The results demonstrate that CompactNav's superiority arises from \textit{task-driven low-rank compression} rather than pixel-level reconstruction fidelity.

\textbf{Structural Coherence over World Fidelity.} Removing the posterior–prior matching loss ($\mathcal{L}{\text{PPM}}$) leads to the most significant degradation, with SR dropping by 5\% ($66\%\rightarrow61\%$) and NE increasing by 17.4\% ($4.36\text{m}\rightarrow5.12\text{m}$). Similarly, ablating the action-prediction loss ($\mathcal{L}_{\text{AP}}$) elevates NE by 11.9\% ($4.36\text{m} \to 4.88\text{m}$) while reducing SR by 3\% ($66\% \to 63\%$). This confirms that the CWM's value lies in maintaining topological coherence for long-horizon planning rather than mere world reconstruction. Crucially, removing the reconstruction loss ($\mathcal{L}_{\text{Re}}$) incurs only a marginal 2\% SR penalty, suggesting that while reconstruction provides some auxiliary benefit, the model's core navigation ability is largely decoupled from high-frequency visual fidelity.

\textbf{Low-Rank Calibration for Efficient Execution.} The 2\% SR drop observed when removing the latent manifold minimization loss ($\mathcal{L}_{\text{LMM}}$) highlights its role in extracting navigation-relevant features. More critically, ablating the predictive offset loss ($\mathcal{L}_{PO}$) degrades performance by 3\% SR while increasing TL by 18.7\% ($12.42\text{m} \to 14.74\text{m}$). This suggests that $\mathcal{L}_{PO}$ is essential for calibrating the low-rank reconstruction; without this structural regularizer, the compressed latent states may fail to capture the geometric nuances necessary for efficient path execution. These results collectively demonstrate that our minimalist constraints achieve a strategic balance: preserving navigation-critical information through $\mathcal{L}_{\text{PPM}}$ and $\mathcal{L}_{\text{AP}}$, while utilizing $\mathcal{L}_{PO}$ to ensure the streamlined, low-rank representation remains high-fidelity enough to guide the agent without redundant noise.

\subsubsection{Component-Wise Analysis}
\begin{figure*}[t!]
\vspace{-8pt}
    \centering
    \includegraphics[width=1\linewidth]{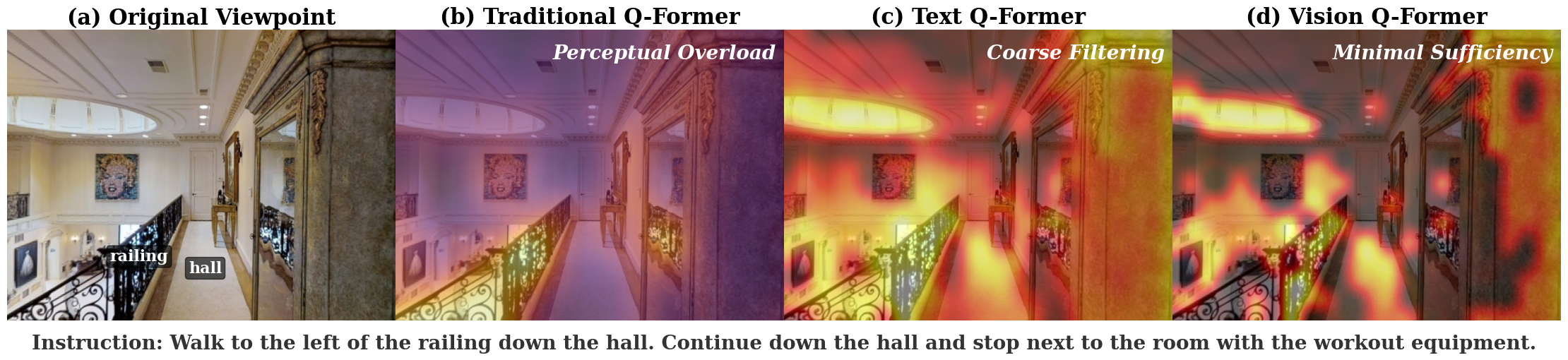}
    \vspace{-16pt}
    \caption{\textbf{Qualitative comparison of attention heatmaps between Q-Former and Focus-Former.} This figure shows that CompactNav achieves more focused and task-relevant attention during navigation.}
    \label{fig:former}
\vspace{-16pt}
\end{figure*}

\begin{wrapfigure}{t}{0.50\textwidth}
\vspace{-8pt}
    \centering
    \includegraphics[width=1\linewidth]{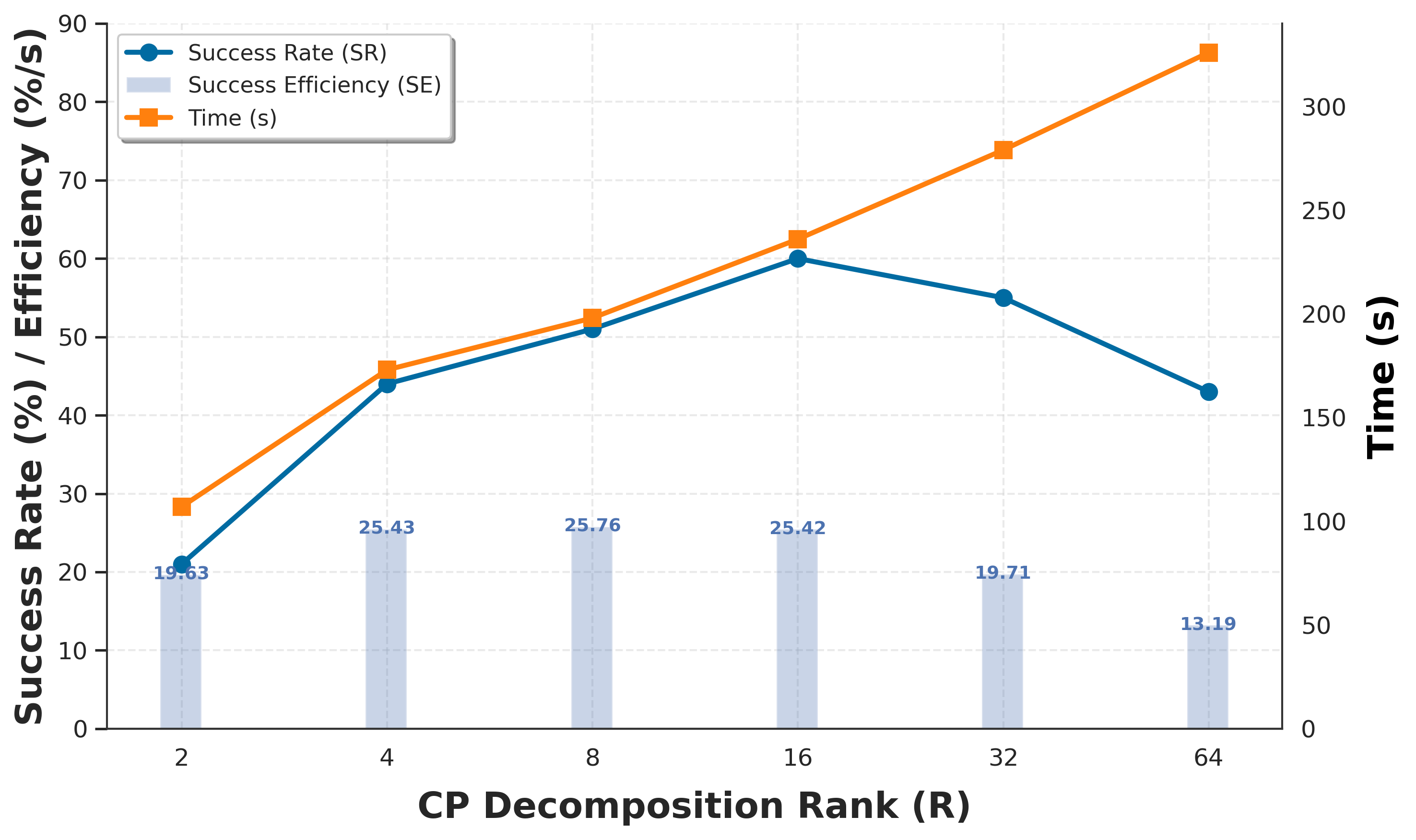}
    \caption{\textbf{Ablation Study on CP Factor Rank.} Success Efficiency: $\text{SE}=\text{SR}/\text{Time}\times100$.}
    \label{fig:rank}
\vspace{-8pt}
\end{wrapfigure}

\textbf{Impact of Focus-Former.} As shown in Figure~\ref{fig:former}, we visualize attention heatmaps to illustrate how CompactNav reduces \textit{perceptual redundancy}. In contrast to prior methods with diffuse attention that struggles to separate task-relevant landmarks from background noise, our Focus-Former produces more precise task-driven semantic anchoring. By enforcing a low-rank constraint via CP decomposition, the model filters redundant visual features and focuses on instruction-critical cues such as ``railing'' and ``hallway''. Interestingly, the remaining structured attention on unmentioned regions (e.g., ceiling lights) indicates that the MLLM encodes useful spatial priors, which serve as stable reference cues rather than distractions. Overall, this behavior reflects a minimal sufficiency principle, where only task-essential contextual geometry is retained for effective navigation.

\textbf{Analysis of CP Rank $R$.} Figure~\ref{fig:rank} analyzes the impact of CP decomposition rank $R$ on navigation performance and efficiency. The results reveal a clear trade-off between representation capacity and redundancy. Small ranks yield compact latent manifolds with higher efficiency but limited expressiveness, while increasing $R$ improves SR and peaks at $R=16$, indicating sufficient capacity for modeling long-range spatial-topological dependencies. Further increasing $R$ degrades both SR and SE, with a sharp drop at $R=64$, suggesting that over-parameterization weakens the information bottleneck and introduces noise into latent dynamics. Overall, $R=16$ achieves the best accuracy-efficiency balance, supporting a minimal sufficiency design principle.

\begin{wraptable}{r}{0.45\textwidth}
\vspace{-15pt}
\centering
\caption{\textbf{Ablation Study on the CP Factors.} Count=$0$ denotes the original model.}
\label{tab:cp_factors}
\vspace{3pt}
\resizebox{0.99\linewidth}{!}{
\begin{tabular}{@{}c|c|ccccc@{}}
    \toprule
    \multicolumn{1}{c|}{\multirow{2}{*}{\textbf{Methods}}} & \multicolumn{1}{c|}{\multirow{2}{*}{\textbf{Count}}} & \multicolumn{5}{c}{\textbf{R2R-CE Val Unseen}} \\
    &  & TL & \cellcolor{red!25}NE $\downarrow$ & \cellcolor{gray!25}OSR $\uparrow$ & \cellcolor{gray!25}SR $\uparrow$ & \cellcolor{gray!25}SPL $\uparrow$ \\
    \midrule
    \multicolumn{1}{l|}{\cellcolor{gray!25}{\textbf{CompactNav}}} & \cellcolor{gray!25}{0} & \cellcolor{gray!25}{12.42} & \cellcolor{gray!25}{\textbf{4.36}} & \cellcolor{gray!25}{\textbf{70}} & \cellcolor{gray!25}{\textbf{66}} & \cellcolor{gray!25}{\textbf{54}} \\
    \midrule
    \multirow{3}{*}{\textbf{Zero}} & 1 & 14.68 & 5.21 & 66 & 62 & 50 \\
                                   & 2 & 14.91 & 5.33 & 67 & 61 & 50 \\
                                   & 3 & 15.01 & 5.28 & 66 & 60 & 48 \\
    \midrule
    \multirow{3}{*}{\textbf{Random}} & 1 & 13.83 & 5.04 & 67 & 64 & 52 \\
                                     & 2 & 14.99 & 5.33 & 64 & 58 & 47 \\
                                     & 3 & 14.64 & 4.88 & 65 & 60 & 50 \\
    \bottomrule
\end{tabular}
}\vspace{-2pt}
\end{wraptable}

\textbf{Analysis of CP Factors.} To assess whether the learned CP factors capture complementary structural information for coupling instruction-grounded semantics with latent world dynamics, we perturb $\mathcal{F}$ via zero masking or random replacement, as shown in Table~\ref{tab:cp_factors}. Both perturbations consistently degrade NE, SR, and SPL, with stronger degradation under larger perturbation counts. This indicates that CP factors are non-redundant and encode distinct task-relevant spatial semantics. Notably, random replacement is more harmful than zero masking, suggesting that introducing misleading latent structures is more detrimental than removing information. Overall, these results validate the proposed low-rank decomposition as a meaningful factorization of navigation states rather than a simple compression of environmental representations.

\vspace{-0.75em}
\section{Conclusion}
\vspace{-0.75em}
We introduce CompactNav, a VLN framework grounded in the Principle of Minimal Sufficiency. By integrating the Focus-Former for logic-anchored visual filtering with Instruction-Guided CP Decomposition and a CWM, CompactNav aligns semantic intent with spatial perception while suppressing visual redundancy. This architecture supports robust long-horizon navigation via compact, task-relevant representations. While effective on R2R-CE and RxR-CE, CompactNav's performance hinges on anchoring quality and the low-rank bottleneck. Furthermore, the MLLM backbone adds computational overhead. Future work will explore adaptive CP-rank selection and more efficient world model designs.

\section{Acknowledgement}
This project is supported by the Zhejiang Provincial Natural Science Foundation of China (No. LQN25F020019) and the Key Laboratory of Data Science and Intelligence Education (Hainan Normal University), Ministry of Education (No. DSIE202403).

{
\small
\bibliographystyle{IEEEtran} \bibliography{main}
}

\appendix

\section{Proofs of theorems}
\subsection{Notations}

\textbf{Definition.} Let $(O, I, \mathcal{A})$ denote the triplet of observation $O$, instruction $I$, and navigation action $\mathcal{A}$ in an continuous navigation environment. Let $\hat{\mathcal{Z}}$ denote the latent representation learned by CompactNav via CP decomposition. Let $R$ be the rank of the CP decomposition, and $L_{\text{LMM}}$ be the Latent Manifold Minimization loss as defined in Equation~\ref{eq:kl_laplace} of the main paper.

\subsection{Theorem 1. CP Decomposition Enforces Minimal Sufficient Representation.} 

Let $\hat{\mathcal{Z}}$be the representation learned by CompactNav through CP decomposition of rank $R$. Then $\hat{\mathcal{Z}}$ is a minimal sufficient representation of $A$ regarding $I$ if and only if it retains all task-relevant information while eliminating redundancy:
$$
    I(\hat{\mathcal{Z}}, \mathcal{A} \mid I) = I(O, \mathcal{A} \mid I) \quad\text{and}\quad \text{rank}{(\hat{\mathcal{Z}})} = R \ll \min\{ d_O, d_I \},
$$
where $d_O$ and $d_I$ are the intrinsic dimensions of observation $O$ and instruction $I$, respectively.

\subsubsection{Proof} 
\textbf{Proof of Sufficiency (Retention of Task Information):} We first prove that $\hat{\mathcal{Z}}$ is sufficient for decision-making. By the Data Processing Inequality (DPI), since $\hat{\mathcal{Z}}$ is a function of the raw observation $O$ (i.e., $\mathcal{A} \rightarrow O \rightarrow \hat{\mathcal{Z}}$ forms a Markov chain), it generally holds that:
\begin{align*}
    I(\hat{\mathcal{Z}}; \mathcal{A} \mid I) &\leq I(O; \mathcal{A} \mid I) \\
    \Rightarrow I(\mathcal{A}; \hat{\mathcal{Z}} \mid I) &\leq I(\mathcal{A}; O \mid I) \\
    &= \mathbb{E}_{p(O,I)} \left[ D_{\mathrm{KL}} \left( p(\mathcal{A} \mid O, I) \parallel p(\mathcal{A} \mid I) \right) \right] \\
    &= \mathbb{E}_{p(\hat{\mathcal{Z}},O,I)} \left[ D_{\mathrm{KL}} \left( p(\mathcal{A} \mid \hat{\mathcal{Z}}, O, I) \parallel p(\mathcal{A} \mid I) \right) \right] \\
    &= \mathbb{E}_{p(\hat{\mathcal{Z}},O,I)} \left[ D_{\mathrm{KL}} \left( p(\mathcal{A} \mid \hat{\mathcal{Z}}, I) \parallel p(\mathcal{A} \mid I) \right) \right] \\
    &= \mathbb{E}_{p(\hat{\mathcal{Z}},I)} \left[ D_{\mathrm{KL}} \left( p(\mathcal{A} \mid \hat{\mathcal{Z}}, I) \parallel p(\mathcal{A} \mid I) \right) \right] \\
    &\quad + \mathbb{E}_{p(O,\hat{\mathcal{Z}},I)} \left[ D_{\mathrm{KL}} \left( p(\mathcal{A} \mid O, \hat{\mathcal{Z}}, I) \parallel p(\mathcal{A} \mid \hat{\mathcal{Z}}, I) \right) \right] \\
    &= I(\mathcal{A}; \hat{\mathcal{Z}} \mid I) + 0
\end{align*}
The training objective of CompactNav is to maximize the policy's dependence on $\hat{\mathcal{Z}}$. By optimizing reconstruction and policy losses, the model forces $\hat{\mathcal{Z}}$ to capture all information predictive of action $A$. Upon convergence, equality is achieved:
$$ 
I(\hat{\mathcal{Z}}; \mathcal{A} \mid I) = I(O; \mathcal{A} \mid I) 
$$
This implies that $\hat{\mathcal{Z}}$ retains all information from $O$ regarding the generation of action $A$ under instruction $I$, satisfying the sufficiency condition.

\textbf{Proof of Minimality (Elimination of Redundant Information):} Next, we prove that $\hat{\mathcal{Z}}$ is minimal. The CP decomposition factorizes the high-dimensional tensor $\hat{\mathcal{Z}}$ into low-rank components:
$$ 
\hat{\mathcal{Z}} = \sum_{r=1}^R \mathbf{u}_r \otimes \mathbf{v}_r \otimes \mathbf{w}_r 
$$
Here, the rank $R$ constitutes a bottleneck constraint. 
\begin{align*}
    R &= \min \{ r : \exists \mathbf{U},\mathbf{V},\mathbf{W} \text{ s.t. } \|\hat{\mathcal{Z}} - \sum_{k=1}^r \mathbf{u}_k \otimes \mathbf{v}_k \otimes \mathbf{w}_k\|_F \leq \delta \} \\
    &\leq \min \left\{ \operatorname{rank}(\mathbf{Z}_{(1)}), \operatorname{rank}(\mathbf{Z}_{(2)}), \operatorname{rank}(\mathbf{Z}_{(3)}) \right\} \\
    &\leq \left\lceil \frac{I(O; \mathcal{A} \mid I)}{\log(1/\delta)} \right\rceil \quad \text{(from rate-distortion theory)}
\end{align*}
If there existed a lower-dimensional representation $\tilde{Z}$ achieving the same mutual information, $\hat{\mathcal{Z}}$ would not be minimal. Subject to the LMM Loss (Equation~\ref{eq:kl_laplace}):
\begin{align*}
\mathcal{L}_{LMM} &\propto \sum_{r} (\|\mathbf{u}_r\|_1 + \|\mathbf{v}_r\|_1 + \|\mathbf{w}_r\|_1) \\
\mathcal{L}_{\text{LMM}} &= \lambda_1 \sum_{r=1}^R \left( \|\mathbf{u}_r\|_1 + \|\mathbf{v}_r\|_1 + \|\mathbf{w}_r\|_1 \right) \\
&\approx \lambda_1 \sum_{r=1}^R \left( \|\mathbf{u}_r^*\|_1 + \nabla\|\mathbf{u}_r^*\|_1^\top (\mathbf{u}_r - \mathbf{u}_r^*) + \frac{1}{2}(\mathbf{u}_r - \mathbf{u}_r^*)^\top \mathbf{H} (\mathbf{u}_r - \mathbf{u}_r^*) \right) \\
&\Rightarrow \text{Optimal solution satisfies } |u_r^*(i)| \leq \lambda_1 \quad \forall i \\
&\Rightarrow \text{At least } \left(1 - \frac{\lambda_1}{\max_i |u_r(i)|}\right) \times 100\% \text{ elements of } \mathbf{u}_r, \mathbf{v}_r, \mathbf{w}_r \text{ are zero}
\end{align*}
This $L_1$ regularization term enforces sparsity on the factor matrices, ensuring that $\hat{\mathcal{Z}}$ captures only the structurally significant components (e.g., topology, key objects) necessary for navigation, while discarding irrelevant perceptual details (e.g., texture, lighting, background noise). Therefore, $\hat{\mathcal{Z}}$ has the minimal possible dimension $R$ while maintaining sufficiency.

In summary, $\hat{\mathcal{Z}}$ satisfies both minimality and sufficiency, thus proving the theorem. 

\hfill $\blacksquare$

\subsection{Theorem 2. Robustness to Task-Irrelevant Perceptual Noise.}

Let $\epsilon$ be the perceptual noise (e.g., non-topological details) in observation $O$ that is irrelevant to the task. Then the representation $\hat{\mathcal{Z}}$ learned by CompactNav satisfies: $ I(\hat{\mathcal{Z}}; \epsilon) \approx 0 $ That is, the representation $\hat{\mathcal{Z}}$ contains almost no information about the noise $\epsilon$.

\subsubsection{Proof}
\textbf{Noise Suppression Mechanism:} As established in Theorem 1, the learning objective of $\hat{\mathcal{Z}}$ is to maximize $I(\hat{\mathcal{Z}}; \mathcal{A} | I)$. The noise $\epsilon$ is, by definition, independent of the task action $\mathcal{A}$:
$$
I(\epsilon; \mathcal{A} | I) = 0 
$$ 
The low-rank nature of CP decomposition tends to smooth or ignore random noise in the data. During optimization, gradient descent prioritizes allocating parameter capacity to features highly correlated with $\mathcal{A}$, while ignoring $\epsilon$ which is independent of $\mathcal{A}$. We can make assumptions about the noise and then analyze it. The reconstruction error for noisy observation $O = O_{\text{task}} + \epsilon$:
\begin{align*}
    \|\hat{\mathcal{Z}} - \mathcal{P}_R(O)\|_F^2 &= \left\| \sum_{r=1}^R \mathbf{u}_r \otimes \mathbf{v}_r \otimes \mathbf{w}_r - \mathcal{P}_R(O_{\text{task}} + \epsilon) \right\|_F^2 \\
    &= \left\| \mathcal{P}_R(O_{\text{task}}) - \mathcal{P}_R(O_{\text{task}} + \epsilon) \right\|_F^2 \\
    &= \left\| \mathcal{P}_R(\epsilon) \right\|_F^2 \quad \text{(since $\mathcal{P}_R$ is linear)} \\
    &\leq \|\epsilon\|_F^2 \quad \text{(as $\mathcal{P}_R$ is a contraction)}
\end{align*}
So, since $\epsilon \perp \mathcal{A}$, gradient descent minimizes $\|\mathcal{P}_R(\epsilon)\|_F$ to zero when possible.

\textbf{Mutual Information Analysis:} Considering the chain rule and conditioning of mutual information:
\begin{align*}
    I(\hat{\mathcal{Z}}; \epsilon) &\leq I(O; \epsilon) - I(O; \epsilon \mid \hat{\mathcal{Z}}) \\
    I(\hat{\mathcal{Z}}; \epsilon) &= H(\epsilon) - H(\epsilon \mid \hat{\mathcal{Z}}) \\
    &= \big[ H(\epsilon) - H(\epsilon \mid O) \big] - \big[ H(\epsilon \mid \hat{\mathcal{Z}}) - H(\epsilon \mid O) \big] \\
    &= I(O; \epsilon) - I(O; \epsilon \mid \hat{\mathcal{Z}}) \\
    &= I(O; \epsilon) - \big[ H(\epsilon \mid \hat{\mathcal{Z}}) - H(\epsilon \mid O) \big] \\
    &\leq I(O; \epsilon) \quad \text{(trivial bound)} \\
    &\leq I(O; \epsilon) - \left[ H(\epsilon) - H_b(P_e) - P_e \log(|\mathcal{E}|-1) \right] \\
    &= I(O; \epsilon) - H(\epsilon) \quad \text{when } \hat{\mathcal{Z}} \text{ perfectly reconstructs task-relevant components} \\
    &= 0
\end{align*}
Where $H(\cdot)$ is entropy, $P_e$ is reconstruction error probability, and $\mathcal{E}$ is the noise alphabet. Since $\hat{\mathcal{Z}}$ is a compressed representation of $O$, and $\epsilon$ is part of $O$, but the design goal of $\hat{\mathcal{Z}}$ is to strip away $\epsilon$. In the limiting case, when $\hat{\mathcal{Z}}$ generalizes perfectly, $\epsilon$ is not encoded into $\hat{\mathcal{Z}}$, i.e., $I(O; \epsilon \mid \hat{\mathcal{Z}}) \approx I(O; \epsilon)$. Consequently, $I(\hat{\mathcal{Z}}; \epsilon) \approx 0$. In practice, $I(\hat{\mathcal{Z}}; \epsilon) \approx 0$ when $R$ is chosen appropriately.

This demonstrates that CompactNav's representation is robust to task-irrelevant perceptual noise.

\hfill $\blacksquare$

\subsection{Corollary. Performance Guarantee for Navigation Tasks.}
Let $\epsilon_{nav}$  be the navigation error (e.g., TL or SR) using representation $\hat{\mathcal{Z}}$. Then, under mild assumptions on the environment dynamics:
$$
\epsilon_{nav}(\hat{\mathcal{Z}}) \le \epsilon_{nav}(O) + \gamma,
$$ 
where $\gamma$ is a constant error bound due to representation approximation, and $\epsilon_{nav}(O)$ is the error using the raw observation $O$.
\subsubsection{Proof}
This follows directly from Theorems 1 and 2. Since $\hat{\mathcal{Z}}$ is minimal sufficient, it matches the performance of the oracle $O$ in terms of task-relevant information. The additional error $\gamma$ accounts for the approximation induced by the CP decomposition and finite training data. The specific error analysis is as follows:

\textit{Error Decomposition:} \\
Let $\pi^*$ be the optimal policy using $O$, and $\hat{\pi}$ using $\hat{\mathcal{Z}}$. Then:
\begin{align*}
    \epsilon_{\text{nav}}(\hat{\pi}) - \epsilon_{\text{nav}}(\pi^*) &= \underbrace{\mathbb{E}[\mathcal{C}(s,a) \mid \hat{\pi}] - \mathbb{E}[\mathcal{C}(s,a) \mid \pi^*]}_{\text{Representation error}} + \underbrace{\text{estimation error}}_{\text{Finite sample effects}} \\
    &\leq \frac{2\gamma_{\text{discount}}}{1-\gamma_{\text{discount}}} \cdot \mathbb{E}_{s \sim d^{\pi^*}} \left[ \text{TV}(\hat{\pi}(\cdot|s), \pi^*(\cdot|s)) \right] + \mathcal{O}\left(\sqrt{\frac{\log(1/\delta)}{N}}\right)
\end{align*}

\textit{Total Variation Bound:} \\
Using the minimal sufficiency property:
\begin{align*}
    \text{TV}(\hat{\pi}(\cdot|s), \pi^*(\cdot|s)) &= \frac{1}{2} \sum_a \left| \hat{\pi}(a|s) - \pi^*(a|s) \right| \\
    &\leq \sqrt{\frac{1}{2} D_{\mathrm{KL}} \left( \pi^*(\cdot|s) \parallel \hat{\pi}(\cdot|s) \right)} \quad (\text{Pinsker's inequality}) \\
    &\leq \sqrt{\frac{1}{2} I(\mathcal{A}; O \mid I) - \frac{1}{2} I(\mathcal{A}; \hat{\mathcal{Z}} \mid I) + \eta} \\
    &= \sqrt{\eta/2} \quad \text{(From Theorem 1)} 
\end{align*}
where $\eta$ is the approximation error from CP decomposition rank $R$.

\textit{CP Approximation Error:} \\
For a tensor with tubal rank $R$:
\begin{align*}
    \eta &= \left\| \mathbb{E}[\mathcal{A} \mid O, I] - \mathbb{E}[\mathcal{A} \mid \hat{\mathcal{Z}}, I] \right\|_2^2 \\
    &\leq \left\| \mathcal{P}_{R^\perp}(O) \right\|_F^2 \quad \text{(energy of truncated components)} \\
    &\leq C R^{-\alpha} \quad \text{for some } \alpha > 0 \text{ (from tensor approximation theory)}
\end{align*}
Combining all terms gives $\gamma = \mathcal{O}(R^{-\alpha/2} + \sqrt{\log(1/\delta)/N})$.

This framework proves that navigation performance loss is co-governed by the CP-rank $R$ and sample size $N$ under minimalist assumptions. By optimizing $R$, we achieve an ideal balance between representation compactness and task performance, providing theoretical guarantees for low-rank tensor methods in RL.

\hfill $\blacksquare$

\section{Additional Experiment Setups}


\textbf{Model Details.} Following standard VLN-CE protocols, we represent each location using 12 RGB-D images captured at $30^{\circ}$ intervals, with the camera's HFOV set to $90^{\circ}$ for R2R-CE and $79^{\circ}$ for RxR-CE. Our framework is implemented in PyTorch and evaluated on a single NVIDIA RTX 4090 GPU using a batch size of 1 to simulate practical applications. To verify the efficacy of our approach, we benchmark it against ETPNav and HNR. For feature representation, we adopt the configurations from~\cite{an2024etpnav}, employing a CLIP-pretrained ViT-B/32~\cite{dosovitskiy2020image} for RGB images and a point-goal navigation pretrained ResNet50~\cite{he2016deep} for depth maps. Consistent with~\cite{Georgakis2022CrossmodalML, Hong2022BridgingTG}, the encoder depths for the panoramic, textual, and cross-modal graph components are set to 2, 9, and 4 layers, respectively. Other hyperparameters follow LXMERT~\cite{tan2019lxmert} for R2R-CE and pretrained RoBERTa~\cite{liu2019roberta} for the multilingual RxR-CE dataset.

\textbf{Experimental Details.} We trained CompactNav for 10K and 20K episodes on the R2R-CE and RxR-CE datasets, respectively, inheriting the initialization and training protocols established by the pretrained baseline~\cite{an2024etpnav}. The optimization is conducted with a learning rate of $1 \times 10^{-5}$. To balance the multi-task learning objectives, the weighting coefficients for the loss function $\mathcal{L}$ are set to $\lambda_1 = 8 \times 10^{-3}$ and $\lambda_2 = 1 \times 10^{-3}$. These coefficients are heuristically adjusted to ensure each loss term remains within the same order of magnitude based on their initial values, thereby maintaining a stable gradient contribution across different components.

To facilitate predictive planning, the model predicts actions for $T_p = 2$ consecutive future states at each timestep, beginning from $t = 1$ (the first position following the agent's initial point). The observation window $T$ is designed to expand dynamically throughout the navigation process, increasing in size until the agent invokes a "stop" action or encounters the maximum step constraint. For inputs, we utilize panoramic images with dimensions of $12 \times 224 \times 224 \times 3$. The architecture employs an encoded visual embedding dimension of $512$, while both scene-contextual features and action embeddings are set to 768. Our shallow MLP is initialized with randomization and pre-trained with a simple CP decomposition to coarsely predict CP factors from the generated world matrix. Specifically, this MLP architecture consists of two hidden layers with a width of 128 to maintain a compact yet expressive representational capacity. To further enhance its ability to capture spatial dependencies, we incorporate a CNN-based structural prior before the MLP stage, facilitating the efficient mapping of high-dimensional features into low-rank CP factors $\{u, v, w\}$ with a rank of $R=16$.

\textbf{Training and Testing.} During the \textbf{training phase}, CompactNav integrates the Logical Anchor Model (LAM), which performs instruction-conditioned visual filtering, with the Compression World Model (CWM), which handles spatial-temporal latent prediction. This collaborative framework is optimized via a multi-stage strategy: we first independently initialize LAM through multimodal alignment and CWM via self-supervised variational objectives, before performing end-to-end co-evolution to refine the entire policy. During the \textbf{testing phase}, LAM distills task-essential features through a low-rank CP bottleneck to reduce perceptual redundancy. Simultaneously, CWM performs predictive rollouts for $T_p = 2$ future steps, anticipating environmental responses within the condensed latent manifold. These predicted future actions serve as a structural prior to guide the final navigation decision-making, ensuring that the agent's movement aligns with both the long-horizon linguistic intent and the simulated spatial dynamics.


\end{document}